# Event Camera as Region Proposal Network

1st Shrutarv Awasthi
*Chair of Materials Handling and Warehousing*
*Technical University Dortmund*
Germany
shrutarv.awasthi@tu-dortmund.de

2nd Anas Gouda
*Chair of Materials Handling and Warehousing*
*Technical University Dortmund*
Germany
anas.gouda@tu-dortmund.de

3rd Richard Julian Lodenkaemper
*Chair of Materials Handling and Warehousing*
*Technical University Dortmund*
Germany

4th Dr. Moritz Roidl
*Chair of Materials Handling and Warehousing*
*Technical University Dortmund*
Germany

*Abstract*—The human eye consists of two types of photoreceptors, rods and cones. Rods are responsible for monochrome vision, and cones for color vision. The number of rods is much higher than the cones, which means that most human vision processing is done in monochrome. An event camera reports the change in pixel intensity and is analogous to rods. Event and color cameras in computer vision are like rods and cones in human vision. Humans can notice objects moving in the peripheral vision (far right and left), but we cannot classify them (think of someone passing by on your far left or far right, this can trigger your attention without knowing who they are). Thus, rods act as a region proposal network (RPN) in human vision. Therefore, an event camera can act as a region proposal network in deep learning Two-stage object detectors in deep learning, such as Mask R-CNN, consist of a backbone for feature extraction and a RPN. Currently, RPN uses the brute force method by trying out all the possible bounding boxes to detect an object. This requires much computation time to generate region proposals making two-stage detectors inconvenient for fast applications. This work replaces the RPN in Mask-RCNN of detectron2 with an event camera for generating proposals for moving objects. Thus, saving time and being computationally less expensive. The proposed approach is faster than the two-stage detectors with comparable accuracy

*Index Terms*—Event camera, clustering, RPN, High speed perception

## I. INTRODUCTION

Rods and cones are two types of photoreceptor cells located in the human eye's retina. They detect light and send signals to the brain to create visual images. Rods are more sensitive to light and are responsible for vision in low-light conditions, such as at night. They are more densely packed in the outer edges of the retina and are not found in the fovea, which is the central part of the retina responsible for sharp, detailed vision. Cones, on the other hand, are responsible for color vision and visual sharpness. They are more densely packed in the fovea and less light-sensitive than rods. There are three types of cones, each responding to different wavelengths of light and enabling us to perceive a wide range of colors. Overall, the combination of rods and cones in the human eye allows us to see a wide range of colors and visual details in different lighting conditions [28, 12, 13]. An event camera is analogous to rods by reporting a stream of asynchronous events, where each event corresponds to the intensity change of a pixel. Event and color cameras in computer vision are like rods and cones in human vision.

Object detection is one of the most researched problems of computer vision. It forms the basis of many computer vision tasks, such as instance segmentation, image captioning, person detection, and object tracking. Some of the commonly used two-stage techniques for object detection are Faster R-CNN [25], Mask R-CNN [9], and Pyramid networks [15]. For Mask R-CNN, the first stage, the region proposal network (RPN), outputs a set of rectangular bounding box object proposals. The second stage extracts feature using ROI pooling for each proposal and perform classification. In parallel, the Mask R-CNN also outputs a binary mask for each ROI in the second stage. After classification, post-processing (Non-maximum suppression) is performed to eliminate duplicate detections, refine the bounding boxes and re-score the boxes based on other objects in the scene [9]. As a result of the complex two-stage pipeline Mask R-CNN is slow and hard to optimize [23].

One-stage detectors such as YOLO, SSD [19], and RetinaNet [16] predict bounding boxes over the images without the region proposal step. One-stage object detectors achieve a higher detection speed than two-stage detectors but have a lower accuracy [23].

Detectron2 is an open-source system that can be used for tasks such as bounding-box detection, instance, and semantic segmentation. Fig. 1 shows a block diagram of the Mask RCNN implementation of detectron2 architecture (referred to as the base detectron2 in this work). It comprises a Feature Pyramid Network (FPN), Regional Proposal Network (RPN), and ROI Head. The FPN extracts multi-scale feature maps with different receptive fields. These feature maps from the input image are given as input to the RPN. RPN utilizes feature maps to generate box proposals (default is 1000) with confidence scores. The proposed boxes are a grid of anchors

Identify applicable funding agency here. If none, delete this.

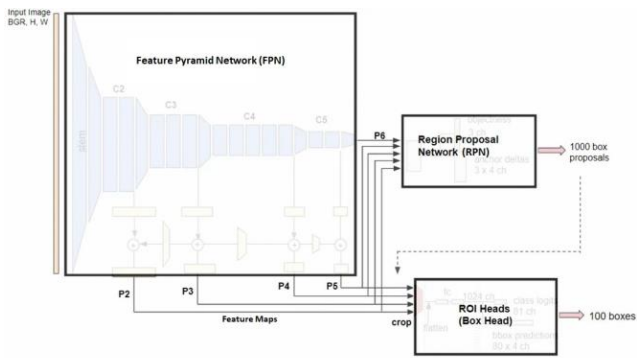

Fig. 1. Block diagram of the Mask RCNN based implementation of detectron2 (referred to as the base detectron2 in this work)[1].

tiled in different aspect ratios and scales. The proposal boxes and feature maps from FPN are given as input to the ROI head. ROI head uses proposal boxes to crop and warp feature maps into multiple fixed-size features. Therefore, the processing time in ROI head is directly proportional to the number of proposal boxes give as input. Finally, fine-tuned bounding box locations and classifications are obtained as the output from ROI head [29].

Conventional frame-based cameras capture redundant frames, increasing power consumption, latency, and bandwidth, especially when tracking moving objects. They also have a fixed dynamic range and struggle to produce clear images under varying illumination [4]. Furthermore, when capturing high-speed motion, the data may have motion blur, large displacements, and occlusions between consecutive frames, causing difficulties in object tracking [3]. On the other hand, event cameras detect fast-moving objects much faster than standard frame-based cameras. They asynchronously output intensity changes for each pixel, computing the difference between current and previous intensity levels and sending an event when the difference is above or below a threshold. This asynchronous approach enhances the dynamic range(140 dB vs. 60 dB of standard cameras [8]) of the event camera and makes it highly useful in low or varying illumination conditions.

An event camera outputs the location (x,y) of the pixel, the timestamp t, and the 1-bit polarity $p$ denoting the change (i.e., intensity increase ("True") or decrease ("False")). The event camera records the dynamic information, not the unnecessary static information such as background landscapes or sky. The output of an event camera is event-driven and frame-less, resulting in low latency, low power consumption, and low bandwidth demands [7] [2]. An event camera generates events when there is motion; therefore, we only consider moving objects for our experiments. We do not move the event camera in this work, as moving it would generate events even for the static background. Differentiating moving objects from the static background in such a case is challenging and will be done as the future step of this work.

Loadrunners [27] [11] and EvoBots [10] are high-speed mobile robots developed by Fraunhofer IML Dortmund and can move up to speeds of 10 m/s. Loadrunners are suitable for warehouse applications such as carrying loads and sorting packages. EvoBots can be used in material handling and also complex urban areas. One of the key challenges faced by high-speed mobile robots is the need for accurate and reliable perception systems. Conventional frame-based cameras are unsuitable for such high-speed perception, as they are limited by their frame rate and can miss critical information when objects move quickly. On the other hand, event cameras are designed to detect changes in the local environment, even when objects are moving at speeds of 20 m/s or higher [5]. Thus event cameras are effective tools for a wide range of high-speed applications.

In this work, we replace the RPN part of the detectron2 with the bounding boxes (BBoxes) generated using the event camera, as seen in Fig. 2. The event camera ignores the static background and only outputs relevant events corresponding to moving objects. The events are clustered and bounding box proposals are created. The BBox proposals are accurate in terms of size and position and are generated faster than the RPN. Furthermore, compared to RPN, the event camera generates fewer BBox proposals. Therefore, for detectron2, the computation time for subsequent steps such as ROI pooling, processing by the head networks, and filtering of BBox also reduces substantially. Moreover, our approach can be used to generate proposals in varying light conditions and for real time applications such as fast moving robots, autonomous driving and high speed drones. To the best of our knowledge, our work is the first attempt to use event camera as a region proposal generator.

RGB and event cameras are fixed on a stand, as shown in 3. A fixed number of events from the event camera are sampled, and a 2$D$ pseudo-image frame matrix (x,y) is created to store the polarity of an event. The sampled events are clustered, and bounding boxes are created for all the valid clusters. These bounding box coordinates are given as input to the ROI head instead of the proposals obtained from the RPN. The COCO evaluator is used for evaluating the bounding boxes. Videos with moving objects are recorded indoors and outdoors. III describes the method in detail. IV explains the dataset, experiments and the results. Finally, the conclusion is presented in V and future works in VI

## II. RELATED WORKS

Clustering is crucial because an accurate and fast clustering algorithm on the event camera data would ensure accurate bounding box creation in real-time. Clustering is an unsupervised machine learning algorithm that works by iteratively partitioning data points into different groups or clusters based on some distance metric [20]. One commonly used distance metric is the Euclidean distance. Some of the popular clustering algorithms include k-means, hierarchical and density-based clustering. DBSCAN, a density-based clustering algorithm, groups data points based on their density in the feature space. Points that are close together and have high density are considered part of the same cluster, while points that are isolated or have low density are considered noise [6].

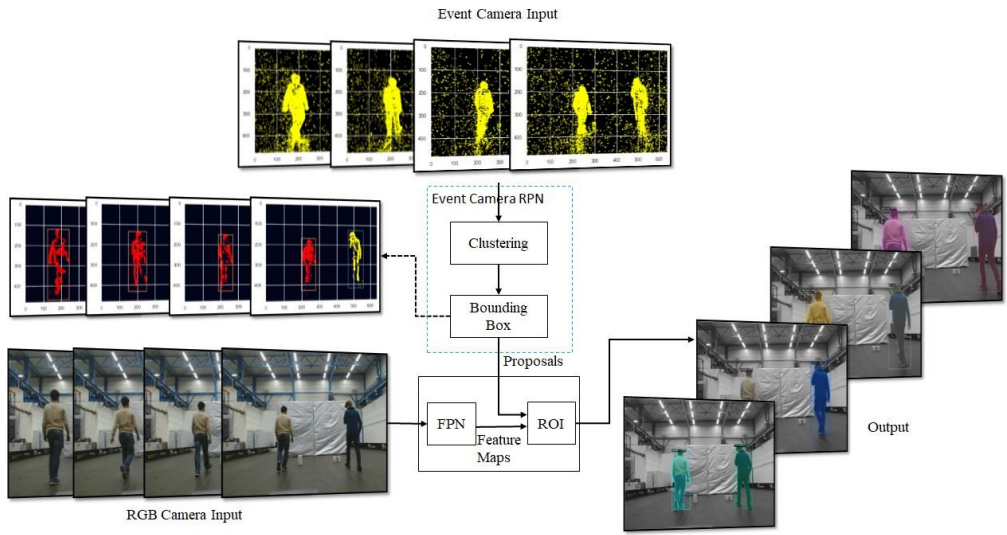

Fig. 2. Architecture of the proposed method. This is referred to as the modified detectron2 architecture in this work.

Mondal et al. present an unsupervised Graph Spectral clustering technique for moving object detection on event-based data. Their method can cluster multiple objects of arbitrary shapes in a scene. They use the silhouette analysis strategy to compute the number of clusters, which is the number of moving objects in a scene [21]. Francisco et al. propose an event-based mean-shift clustering method for real-time event-based clustering and tracking using a Dynamic Vision Sensor. They process each event asynchronously and validate their method for tracking multiple objects. They cluster event-wise and use Kalman filters to reduce computational complexity. They achieved an F-measure of 0.95 for clustering, and compared to the frame-based method, the computational cost was reduced by 88% [3].

YOLO is a deep-learning-based object detection and tracking algorithm that looks at the scene only once. In YOLO, the input image is divided into grid cells. Each cell predicts bounding boxes, the class probability of the object, and a confidence score for each box. YOLO uses a single neural network to simultaneously predict bounding boxes and class probabilities for objects in an image. It is a single-stage detector and is faster than two-stage detectors such as Faster R-CNN and Mask R-CNN. The accuracy of two-stage detectors is typically higher than one-stage detectors [18, 14, 23]. Detectron2 contains a Mask R-CNN but is fast, flexible, and a modular object detection and segmentation tool. It is also a two-stage detector with a backbone network, RPN, and ROI heads. The ROI head contains proposal box sampling, ROI pooling, box head, and mask head. It also filters out the low score, and overlapping bounding boxes [29]. The RPN in two-stage detectors requires much computation to generate region proposals, making the detectors slow and hard to optimize [23]. Therefore, this work aims to replace the RPN in the Mask-RCNN of a detectron2 with the proposals generated from an event camera. As a result, the number of proposals and computation time are reduced.

Camera calibration is an important step when dealing with frames or events. Muglikar et al. propose a framework for calibrating an event camera. The proposed approach can use standard calibration patterns such as a checkerboard pattern. Firstly, the authors divide the recorded event data into chunks of constant time duration. Subsequently, they perform image reconstruction using a neural network. Finally, the kalibr calibration toolbox can be used for intrinsic calibration [24]. An extrinsic calibration between a frame-based and event-based camera can also be performed [22].

III. METHOD

In this section, firstly the hardware setup used in this work is explained. Subsequently the architecture deployed in this work is explained. Next, the algorithm used for clustering event camera data is discussed. Finally, the integration of the clustering algorithm with the detectron2 architecture is explained.

A. Hardware setup

We use a DVxplorer event camera with a resolution of 640x480p and a RGB Logitech C920 HD pro webcam with a resolution of 1080x720p. We fixed both the cameras next to each other on a mount as shown in 3. The distance between the center of the lenses of the two cameras in x,y and z direction is 9, 4, and 3 cm approximately.

We use E2Calib [22] and kalibr [24] for calibrating the event camera. Robot operating system (ROS) is used to record events and images in a rosbag. The recorded events are reconstructed into images using python. Then, the event camera is calibrated using the scripts from the kalibr toolbox. After calibrating the event camera, we tried to calibrate the multi-camera setup using kalibr, but the optimization process during the calibration failed to converge. Therefore, we manually adjust the cameras every time before recording a video to have the maximum field of view (FOV) overlap between the two cameras. We use a windows machine with i5 quad-core processor and 8 Gb of RAM to execute the python scripts.

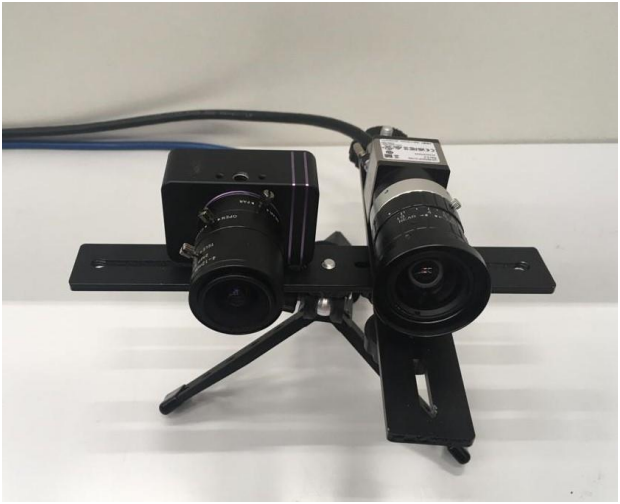

Fig. 3. Our camera setup. Left is a DVXplorer event camera and right a monocular color camera.

### B. Architecture

The architecture used in this work is shown in Fig. 2. The RPN part in the base detectron2, shown in Fig. 1 is replaced with the event camera for generating the region proposals. An RGB image from a camera is gven as input to the FPN. The FPN extracts multi scale feature maps. The event camera based RPN generates region proposal which along with the feature maps are given as input to the ROI Head. The proposals generated are limited to the number of moving objects in the scene. Therefore, compared to base detectron2 the ROI Head in the modified architecture has fewer proposals to process before computing the final bounding boxes. Thus, reducing the computation time. The output from the ROI Head are bounding boxes which are superimposed on the input RGB image for better visualization.

### C. Clustering

The dual camera setup is used to record videos, and each recorded rgb frame has a corresponding event message from the event camera. The recording is done at 30 frames per second (fps), and experiments are performed at 15 fps to facilitate computationally efficient clustering. Each event message has numerous events, and the number of events is directly proportional to the movement in the scene. Events from 10 such sequential event messages are accumulated to form a chunk. The fps for experimentation is half compared to the fps for recording; therefore, events from every second message from the chunk are deleted. This deletion speeds up the clustering without reducing the accuracy.

The events generated due to the moving objects consist of the x and y coordinates of the pixel where an event happened, the timestamp, and the polarity of the event. Polarity is 1 when the pixel brightness increases and 0 when it decreases below a threshold. The polarity is multiplied by a factor of 254 to visualize the events in python. All the events in a chunk are stored in a matrix of size 640x480. The recent events override the old ones. On visualizing the events, one can observe multiple noisy events around moving objects and also in small isolated groups (noise patches) at random places in the image frame. Erosion removes the noise from the event data [26]. In this work, graph-spectral [21] and DBScan[6] clustering is performed to detect moving objects. Clustering is performed sequentially on each chunk of noise-free data. The number of clusters created equals the number of moving objects in the frame. Graph spectral clustering was much slower than DBScan and had comparable accuracy for our datasets. Therefore, we did only present results from DBScan clustering.

Bounding boxes are created around each cluster obtained in the previous step. The coordinates of the bounding boxes are the proposals, which were previously being obtained from the RPN. Proposal generation using event camera is faster than the RPN and can be used in real time for detecting fast moving objects. The proposals are then forwarded to the ROI Heads which outputs the final bounding boxes. The event based clustering algorithm is integrated in the detectron2 architecture, replacing the RPN.

Subsequently, every tenth image is given as input to the FPN of the modified detectron2 architecture. The clustering and bounding box creation using event camera data runs in parallel to FPN, and does not have to wait for an output from the FPN. This will enable parallel processing and reduce total execution time compared to the base detectron2.

We used the COCO evaluator and considered mean average precision (mAP) as the metric. In this work, the object detection evaluation metric must consider both the class and location of the objects. Therefore, mean average precision fits the task. Intersection over union (IoU) is computed for the predicted and the ground truth bounding boxes. We considered the IoU to be higher than 0.75. Additionally the execution time between the base detectron2 and the modified architecture is also computed.

## IV. EXPERIMENTS

In this section the process of recording the videos, the experiments performed and the results are discussed.

### A. Dataset

The cameras were stationary while recording videos in the research hall and outside. The videos recorded inside comprised of two persons performing a pick and place task and a person moving a bottle. The video recorded outside comprised of cars moving on the street. We tested our approach on a total of six videos. Before recording the videos, we adjusted the focus of the event camera to have the maximum overlap between the FOV of the two cameras. The length of each recorded video is between 10–25 secs. 4 shows a frame from each of the recorded video. The videos are annotated using the coco annotator tool [17]. The recorded videos are input to the base detectron2 and the modified architecture shown in 2.

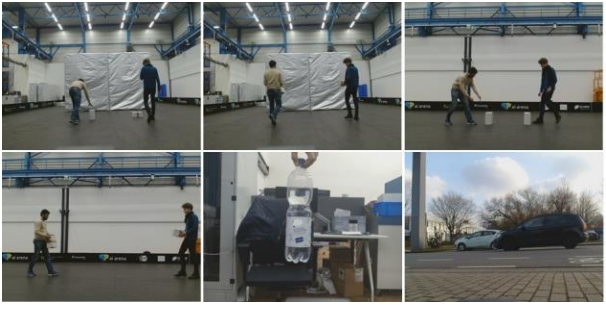

Fig. 4. A frame from each recorded video

*B. Results*

Average precision is calculated for each video using the COCO evaluator on the base detectron2 ($AP_{base}$) and the modified detectron2 ($AP_{mod}$). The bounded boxes generated after clustering are compared with the ground truth bounding boxes, and the average precision is computed. The mean average precision is calculated by taking the mean of all the average precision, calculated previously. Observing the results in I, one can infer that the bounding box proposals generated for moving objects are pretty accurate. The publicly available implementation of DBSCAN and mean shift clustering algorithms when run on a GPU do not significantly improve the performance. Therefore, comparison based on time of our approach with the mask RCNN implementation of detectron 2 is not useful. In order to use GPUs efficiently for clustering we will be using CNNs for clustering as the future work.

TABLE I
AVERAGE PRECISION(AP) VALUES FOR EXPERIMENTS CONDUCTED ON RECORDED VIDEOS. INTERSECTION OVER UNION (IoU) IS CONSIDERED TO BE LARGER THAN OR EQUAL TO 0.75. MEAN AVERAGE PRECISION(mAP) IS CALCULATED BY TAKING THE MEAN OF ALL THE AP. AR IS THE AVERAGE RECALL

| Videos | IoU | #Moving Obj | $AR_{mod}$ | $AR_{base}$ | $AP_{mod}$ | $AP_{orig}$ |
|---|---|---|---|---|---|---|
| Hall 1 | ⩾ 0.75 | 2 | 90.04 | 95.6 | 90.1 | 94.9 |
| Hall 2 | ⩾ 0.75 | 2 | 88.7 | 90.8 | 84.71 | 96.90 |
| Hall 3 | ⩾ 0.75 | 2 | 90 | 92.5 | 92.1 | 95.90 |
| Hall 4 | ⩾ 0.75 | 2 | 97.3 | 94.5 | 90.5 | 91.90 |
| Bottle | ⩾ 0.75 | 1 | 74 | 80.5 | 85.7 | 91.90 |
| Car 1 | ⩾ 0.75 | 1 | 91.2 | 92.5 | 81.2 | 86.1 |
|  |  |  |  |  | mAP = 87.39 | mAP = 92.78 |

## V. Conclusion

Detectron2 uses a backbone network (FPN), RPN, and ROI head. RPN generates 1000 proposals and is computationally expensive. Thus, RPN acts as a bottleneck in detectron2. This work replaces the RPN in the detectron2 with an event camera, as shown in Fig.2. The proposed method uses a dual camera setup to record videos, where each recorded RGB frame has a corresponding event message from the event camera. Clustering is performed on accumulated events to detect moving objects, and bounding boxes are created around each cluster to obtain region proposals. The event camera based region proposal network is integrated into the detectron2 architecture, replacing the RPN, and runs in parallel with the FPN. The proposed method is evaluated using the COCO evaluator on bounding boxes and compared with the base detectron2. We obtained a mean average precision of 0.8739, 5.8% lower than the base detectron2. Overall, event cameras can successfully generate an accurate and exact number of proposals for moving objects. Thus, event cameras can be used as a proposal generator for moving objects. The code for this work can be found in the github repository[1].

## VI. Future Work

Our future work is to replace the classical clustering with a CNN to make the proposals from the event data and extract stationary and moving objects. This has the potential to handle moving clutter better than RGB camera. We also plan to test this setup when the event camera is moving. We have to use the SLAM algorithms and extract moving objects. Moreover, we plan to use a good quality, small RGB camera that can be calibrated with the event camera. A better calibration would increase the proposals' accuracy, giving a higher precision.

## VII. Acknowledgment

The work in this publication was supported the German Federal Ministry of Education and Research (BMBF) in the context of the project "LAMARR Institute for Machine Learning and Artificial Intelligence" (Funding Code: LAMARR22B)

[1]We will give the link to the github repository after our work is published